\documentclass[a4paper,twosided,conference]{IEEEtran}
\def\IEEEkeywordsname{Keywords}

\ifCLASSINFOpdf
\else
\fi

\usepackage{graphicx}
\usepackage[utf8]{inputenc}
\usepackage{amsmath}

\usepackage{amsthm}

\begin{document}
%
% paper title
% can use linebreaks \\ within to get better formatting as desired
\title{OntoRich - A Support Tool for Semi-Automatic Ontology Enrichment and Evaluation}

% author names and affiliations
% use a multiple column layout for up to three different
% affiliations
\author{\IEEEauthorblockN{Adrian Groza\IEEEauthorrefmark{1},
Gabriel Barbur\IEEEauthorrefmark{1},
Bogdan Blaga\IEEEauthorrefmark{1}}, 
%Montgomery Scott\IEEEauthorrefmark{3} and
%Eldon Tyrell\IEEEauthorrefmark{4}}
\IEEEauthorblockA{\IEEEauthorrefmark{1}Computer Science Department\\
Technical University of Cluj-Napoca, Romania\\}
adrian.groza@cs.utcluj.ro\\
}
\maketitle

\begin{abstract}
%\boldmath
This paper presents the OntoRich framework, a support tool for semi-automatic ontology enrichment and evaluation.
The WordNet is used to extract candidates for dynamic ontology enrichment from RSS streams. 
With the integration of OpenNLP the system gains access to syntactic analysis of the RSS news.
The enriched ontologies are evaluated against several qualitative metrics.

\end{abstract}

\IEEEpeerreviewmaketitle

% IEEEtran.cls defaults to using nonbold math in the Abstract.
% This preserves the distinction between vectors and scalars. However,
% if the conference you are submitting to favors bold math in the abstract,
% then you can use LaTeX's standard command \boldmath at the very start
% of the abstract to achieve this. Many IEEE journals/conferences frown on
% math in the abstract anyway.

\IEEEpeerreviewmaketitle

\newtheorem{name}{Printed output}
\newtheorem{defTemp}{Definition}

\section{Introduction}

In recent years, much effort has been put in ontology learning as an imperative for the concept of Semantic Web. 
The migration from Web 2.0 to Semantic Web~\cite{DBLP:journals/dss/DuLK09} is still considered only a theoretical approach mainly because of the effort that this transformation would imply. 
Many solutions were proposed during the recent years both for populating and evaluating ontologies, but working with ontologies is not a straightforward process because some important problems arise. 
First of all, the knowledge needed for populating ontologies is spread over the internet in an unstructured way and information extraction tools have to be designed for each website in particular. 
Information Extraction methods by means of domain specific templates and the lightweight use of Natural Languages Processing techniques (NLP) have been already proposed~\cite{Vargas-VeraDKMB01, Liu2011163}. 
Another good heuristic is to use a search engine to find web pages with relevant content. 
However, current search engines retrieve web pages, not the information itself~\cite{GeleijnseK07}. 
After the information is retrieved, a system for term extraction is needed in order to obtain candidates for ontology enrichment. 
An ontology has to be evaluated against several metrics in order to be considered as a valid ontology for the domain it covers.

The life-cycle of ontologies in the space of Semantic Web involves different techniques, ranging from manual to automatic building, refinement, merging, mapping or annotation. Each technique involves the specification of core concepts for the population
of an ontology, or for its annotation, manipulation, or management~\cite{Gangemi05}. 
These core concepts are referred to as Ontology Design Patterns and represent an important guideline~\cite{GeorgiuGroza11} for the design of an ontology engineering tool, such as the OntoRich system. 
Ontology engineering has become an important domain since the idea of Semantic Web was taken into consideration. It involves various tasks such as editing, evolving, versioning, mapping, alignment, merging, reusing and extraction . The management of available web knowledge is a difficult task because of the dynamic nature of the Internet \cite{art11}. The first consideration was to provide an automatic way for information extraction from the web and the considered solution is based on RSS feeds that more and more websites provide nowadays. An RSS feed provides a standardized XML file format that allows the information to be published once and viewed by many different programs. Because of the standard format a single RSS Reader system is enough to fetch information from many websites that are related to a certain domain.

Ontologies provide explicit formalization and specification of a domain in the form of concepts, their corresponding
relationships and specific instances~\cite{Patel03}. 
The instances contain the actual data that is queried in knowledge based applications. 
Several approaches for extracting concepts, instances and relationships exploit separately or integrate statistical methods, semantic repositories such as WordNet, natural language processing libraries such as OpenNLP, or lexicon-syntactic patterns in form of regular expressions~\cite{art12}.   
The developed system provide users with the capability to choose among and mix these methods in order to obtain potential candidates for ontology enrichment.

Ontology evaluation is an important task in real life scenarious. 
When creating an application based on semantic knowledge it is necessary to guarantee that the considered ontology meets the application requirements. 
Ontology evaluation is also important in cases where the ontology is automatically populated from different resources that might not be homogeneous, leading to duplicate instances, or instances that are clustered according to their sources in the same ontology~\cite{art8}. 
In this line, an important problem is to compare several ontologies that describe the same domain and choose the one that best fits a certain user needs~\cite{art13}. 
%The ontology evaluation has been used nowadays in different applications to find a suitable ontology. 
However, the ontology evaluation is still a challenging task within the semantic web, and especially of ontology engineering. 
The difficulty in choosing one ontology from a number of similar ones is given by the numerous ways you can classify such a structure. 
Due to the fact that an ontology represents a large number of concepts, one can split them in a very large number of ways and categories. 
For example, one can classify ontologies by the abstractness or concreteness of there meaning how good they cover a subject, or how well can they be used in more different subjects~\cite{art14}. 
Moreover, one can split them by the number or relations a given ontology has, or by the way these relations are used between different concepts. 
%Although ontology evaluation techniques are improving and more and more new and innovative measures of evaluation are proposed, the literature contains very few specific examples of explicit evaluation of ontologies, and methods of linking the evaluation process to the development and engineering.

Contributions: This research is an extended version of~\cite{BarburBlagaGroza11}. 
Given the lack of systems designed to manage rapidly changing information at the semantic level~\cite{Valle2009}, 
RSS streams are exploited to extract candidates for dynamic ontology enrichment. 
With the integration of OpenNLP and WordNet the system gains access to syntactic analysis of the RSS streams.

Organisation: 
Section~\ref{sec:architecture} introduces the top level architecture of the system and describes the role of each component.
Section~\ref{sec:capabilities} details the capabilities of the system regarding three vectors: ontology engineering, ontology enrichment, and ontology evaluation. 
Section~\ref{sec:related}  compares the system with existing technical instrumentation, whilst section~\ref{sec:conclusions} concludes the paper.

\section{System Architecture}
\label{sec:architecture}
\begin{figure}[h]
\begin{center}
% Requires \usepackage{graphicx}
\includegraphics[width=3.4in]{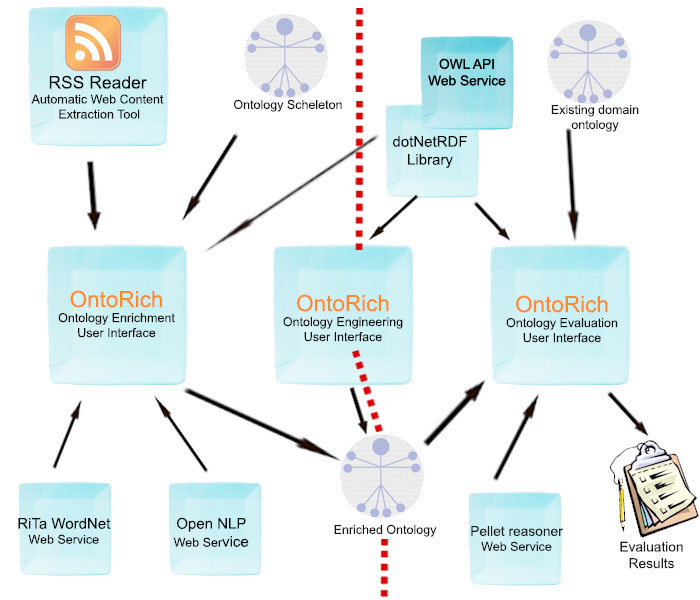}
%\emph{Figure 1: }
\end{center}
\caption{OntoRich system architecture.}
\label{fig:SystemArchitecture1}
\end{figure}

Dealing with ontology population and evaluation involves an engineering process needed for reading and obtaining information from the considered ontology. 
The proposed OntoRich method for ontology engineering is based on dotNetRDF, an Open Source .Net Library using the latest versions of .Net Framework to provide a powerful and easy to use API for working with Resource Description Framework (RDF).
The standard data model RDF extends the linking structure of the Web to use URIs to name the relationship between things as well as the two ends of the link, usually referred to as a $triple$. 
Using this simple model, it allows structured and semi-structured data to be mixed, exposed, and shared across different applications. 
%Our system works with ontologies in the RDF/OWL format.
The main components of the system are the $RSS\ Reader$, the $Ontology\ Engineering$ component, the $Ontology\ Enrichment$ component and the $Ontology\ Evaluation$ module (see figure~\ref{fig:SystemArchitecture1}).

The $RSS\ Reader$ is a web application created in PHP that distinguishes between two main users: the administrator and the normal user. 
The administrator responsibility is to create domains of interest and populate each domain with corresponding RSS feeds. 
A user that enters the site and creates an account has the option of subscribing to one or more domains and receive daily updates by e-mail with content related to the domain of interest. 
An advantage of using RSS is that the information provided is always updated, so new concepts or instances that appear in a domain and are useful to be considered for the managed ontology can be found faster.

The $Ontology\ Engineering$ component is the one dealing with loading, displaying, editing and saving ontologies. 
It is based on the dotNetRDF open source API. dotNetRDF is a .Net library written in C\# designed to provide a simple but powerful API for working with RDF data. 
It provides a large variety of classes for performing all the common tasks from reading and writing RDF data to querying over it. 
The library is designed as highly extensible and allows users to add in support for additional features.

The $Ontology\ Enrichment$ module deals with extracting new terms that can be added as concepts, instances or relations to the ontology. 
It is based on RiTa WordNet Java API and OpenNLP Java API. 
Because the OntoRich system is created using C\# and WPF framework, two web services are needed in order to integrate RiTA WordNet and OpenNLP that are only available in the form of Java API. 
RiTa WordNet is a WordNet library that offers a simple access to the WordNet ontology and also provides distance metrics between ontology terms. 
OpenNLP is an organizational center for open source projects related to natural language processing. 
Its primary role is to encourage and facilitate the collaboration of researchers and developers on such projects. 
OpenNLP also hosts a variety of Java-based NLP tools which perform sentence detection, tokenization, pos-tagging, chunking and parsing, named-entity detection, and co-reference using the OpenNLP Maxent machine learning package.

The $Ontology Evaluation$ provides to users the option of testing the loaded ontology against some defined ontology metrics and also offers some interesting features such as assessing the evolution in time of an ontology, comparing two ontologies or checking an ontology consistency using the Pellet reasoner. 
The major approaches currently in use for the evaluation and validation of ontologies using metric-based ontology quality analysis are available. 
Pellet is an OWL reasoner that provides standard and cutting-edge reasoning services for OWL ontologies. 
It incorporates optimizations for nominals, conjunctive query answering, and incremental reasoning.

\begin{figure}[h]
\begin{center}
% Requires \usepackage{graphicx}
\includegraphics[width=3.4in]{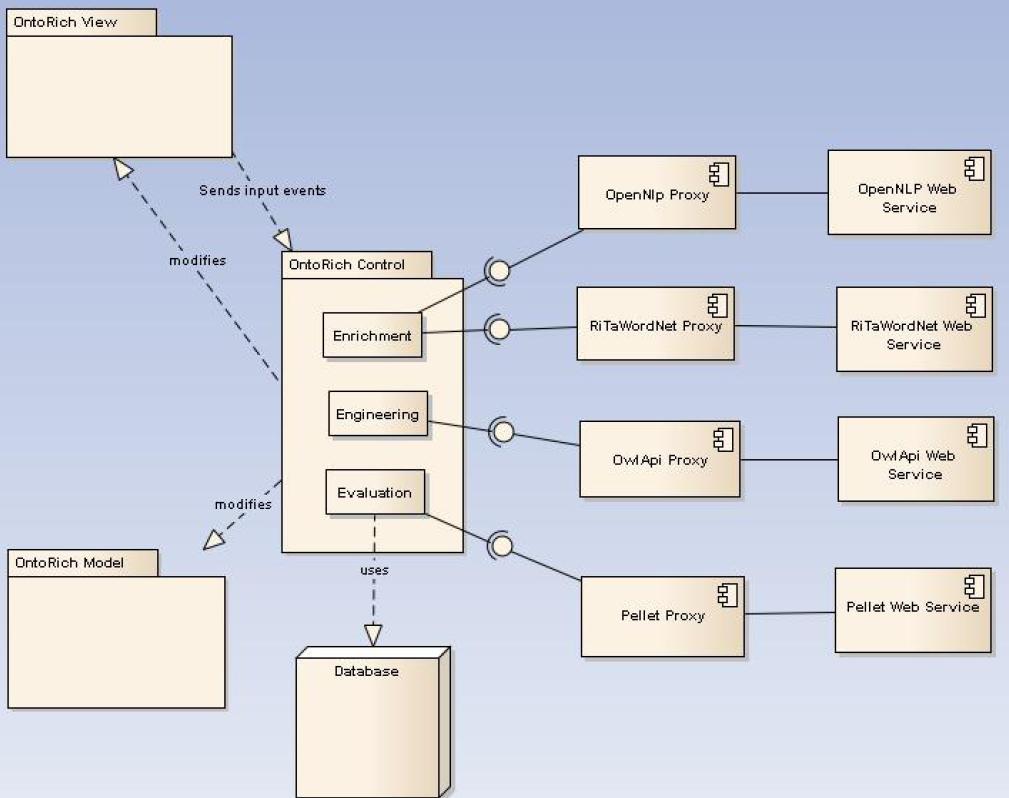}
%\emph{Figure 1: }
\end{center}
\caption{OntoRich component diagram.}
\label{fig:proxy}
\end{figure}

The diagram in figure~\ref{fig:proxy} presents the high level interaction between the OntoRich components and illustrates the implementation of the \textit{proxy design pattern} as a solution for Web services access. 

\section{Framework Capabilities}
\label{sec:capabilities}
The main features of the OntoRich tool\footnote{The sytems is available at http://cs-gw.utcluj.ro/$\sim$adrian/ontorich} are illustrated with the help of two testing ontologies: the well-known 'Wine' ontology and an IT ontology skeleton created using Prot\'eg\'e. 

\subsection{Ontology engineering}

\begin{figure}
\begin{center}
% Requires \usepackage{graphicx}
\includegraphics[width=3.4in]{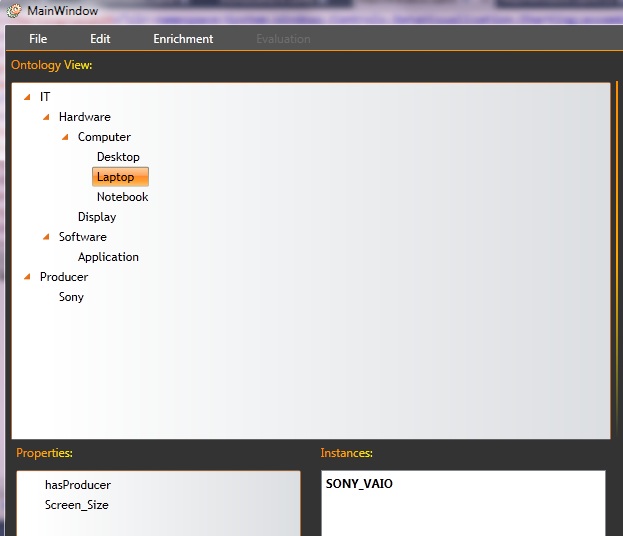}
%\emph{Figure 2: Ontology display}
\end{center}
\caption{Ontology display.}
\label{fig:MainWindow1}
\end{figure}

This section details features related to the management of an ontology. 
%In computer science and information science, an ontology is a formal representation of knowledge as a set of concepts within a domain, and the relationships between those concepts. It is used to reason about the entities within that domain and may be used to describe the domain.
In order to graphically display the ontology, a tree structure is used with nodes representing classes. 
The 'subClassOf' relationship specified in every ontology representation language is used in order to parse the ontology and extract it as a tree view with parent nodes and children nodes. 
The instances of every class can be seen in a separate window as well as the relationships defined in the schema.
The main features that the ontology engineering component provides are: 
i) loading ontologies from a local file or URI;
ii) displaying ontologies in the form of a tree view or in the RDF/OWL format;
iii) displaying ontology relationships and instances in separate windows;
iv) adding concepts, roles, and instances to the ontology;
v) saving the ontology to a specified location.
An example of an ontology display can be seen in figure~\ref{fig:MainWindow1}.

\subsection{Ontology Enrichment}
\label{subsec:merging}
As already mentioned, the Ontology Enrichment tool uses domain categorized web content extracted by our RSS Reader and sent to the user in the form of an e-mail. The e-mail content can be copied in a text corpus within the application. 
Any other text file can be loaded into the corpus and the user can also edit and add text according to its own needs. 
After having a document (or more) added in the corpus the user has several methods for text processing and term extraction.
The first category of term extraction methods is based on two statistical methods absolute term frequency and TF-IDF weight.

\begin{figure}
\begin{center}
% Requires \usepackage{graphicx}
\includegraphics[width=3.4in]{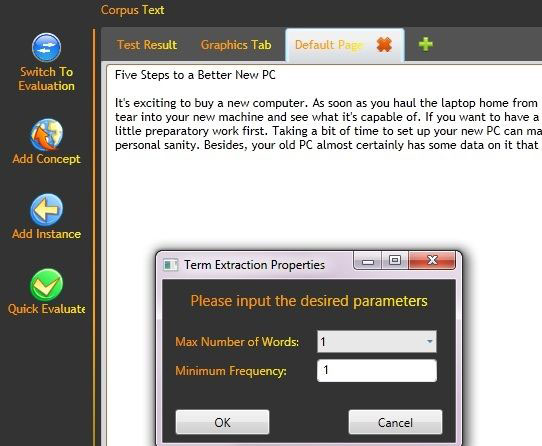}
\end{center}
%\emph{Figure 3: Term Extraction}
\caption{Term extraction.}
\label{fig:TermExtraction1}
\end{figure}

\begin{defTemp}Absolute term frequency $tf_i$ is defined by 
\[
tf_i = \frac{n_i}{\sum_{i=1} n_i}  
 \]
where $n_i$ represents the number of times term $i$ appears.
\end{defTemp}
The system provides options to select the minimum frequency to be considered as well as the maximum number of word in a term (see figure~\ref{fig:TermExtraction1}).

\begin{defTemp}
Term frequency - inverse document frequency metric (TF-IDF weight) evaluates how important a word is to a document in a collection or corpus, defined by:
    \[
    (tf-idf)_{i,j} = tf_{i,j} * idf_i 
    \]
where $tf_{i,j}$ are the absolute term frequency of term $i$ in document $j$ and and $idf_i$ the inverse document frequency, given by
    \[
    idf_i = \log{\frac{|D|}{j:t_i \in d_j}}
    \]
    where $|D|$ is the total number of documents in the corpus and
   $j: t_i \in d_j$ the number of documents where term $t_i$ appears.
\end{defTemp}
The importance increases proportionally to the number of times a word appears in the document but is offset by the frequency of the word in the corpus.

\begin{figure}
\begin{center}
% Requires \usepackage{graphicx}
\includegraphics[width=3.4in]{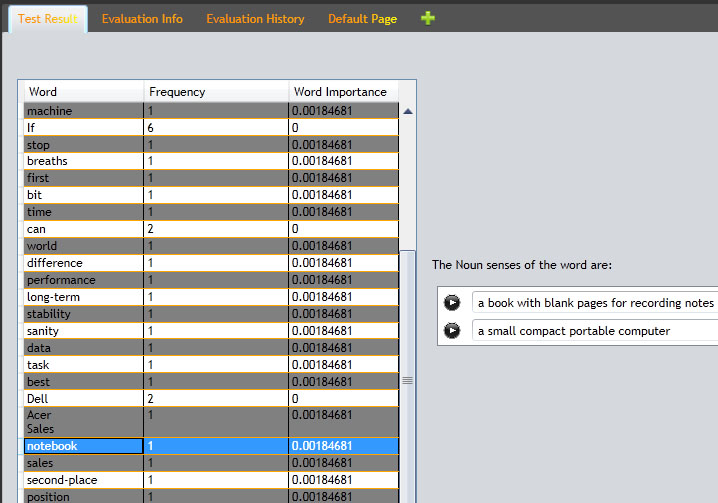}
\end{center}
%\emph{Figure 4: Term Extraction results}
\caption{Term extraction results.}
\label{fig:TermExtractionResult1}
\end{figure}

Using the stemming function provided by RiTa WordNet, each word in the text is reduced to its stem form. 
A word has a single stem, namely the part of the word that is common to all its inflected variants. 
Thus, all derivational affixes are part of the stem. For example, the stem of 'friendships' is 'friendship', to which the inflectional suffix '-s' is attached. Using this approach many forms of basically the same word can be found and counted in computing the statistical values (see figure~\ref{fig:TermExtractionResult1}).

\begin{figure}[h]
\begin{center}
% Requires \usepackage{graphicx}
\includegraphics[width=3.4in]{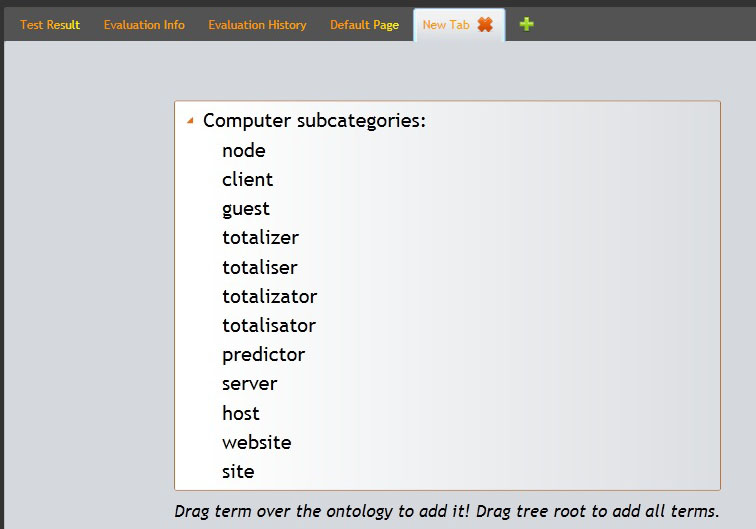}
\end{center}
%\emph{Figure 5: Example of 'partOf' relationship extraction}
\caption{Example of 'partOf' relationship extraction.}
\label{fig:ExtractPartOfRelation1}
\end{figure}

Another feature provided by the OntoRich enrichment component is the possibility of using the existing concepts in the ontology together with the semantic power of WordNet in order to extracting 'partOf', 'membeOf', 'madeFrom' and 'isKindOf' relations. This is made possible by using the methods for retrieving hyponyms and meronyms that RitaWordNet provides. In linguistics, a hyponym is a word or phrase whose semantic field is included within that of another word. For example, 'scarlet', 'vermilion', 'carmine', and 'crimson' are all hyponyms of red (their hypernym), which is, in turn, a hyponym of 'color'. In many ways, meronymy is significantly more complicated than hyponymy. The Wordnet databases specify three types of meronym relationships:
\begin{itemize}
\item Part meronym: a 'processor' is part of a 'computer' (see figure ~\ref{fig:ExtractPartOfRelation1});
\item Member meronym: a 'computer' is a member of a 'computer network';
\item Substance (stuff) meronym: a 'keyboard' is made from 'plastic';
\end{itemize}

\begin{figure}[h]
\begin{center}
% Requires \usepackage{graphicx}
\includegraphics[width=3.4in]{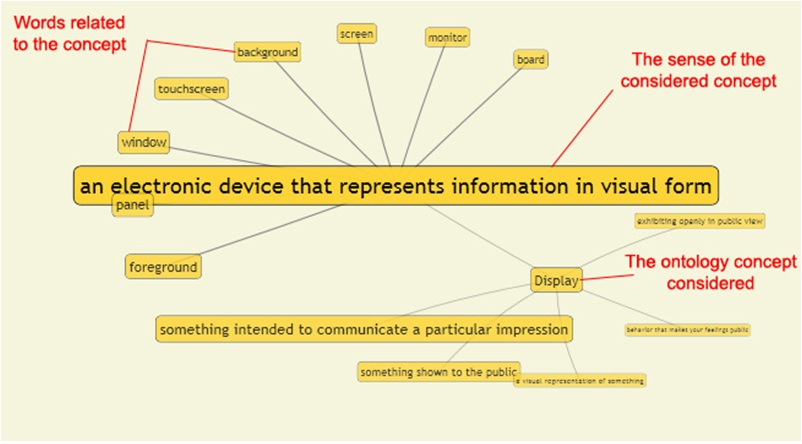}
\end{center}
\caption{Hyponym Tree for the term 'computer'.}
\label{fig:HyponymTreeComputer1}
\end{figure}

More terms can be obtained by using the hyponym tree provided by WordNet to which RiTa WordNet offers a simple access. After selecting a term in the existing ontology the user can display graphically the semantic hierarchy of the word (the hyponym tree rooted at that word). Every word displayed in the hyponym tree can be selected and added to the ontology as child (sub-class) of a specified concept. Results for the IT considered ontology are shown in figure ~\ref{fig:HyponymTreeComputer1}.

In many cases the text corpus could be easier to use if a syntactic analysis could be applied. With the use of the OpenNLP library the OntoRich system provides users the possibility to:

\begin{itemize}
\item split the text into sentences;
\item tag each word with the correct POS(part of speech) within the sentence;
\item use OpenNLP built-in models to extract well-known organization names, person names and date references(e.g. today, Monday, July, etc)(see figure ~\ref{fig:InstanceExtraction1});

\begin{figure}[h]
\begin{center}
% Requires \usepackage{graphicx}
\includegraphics[width=3.4in]{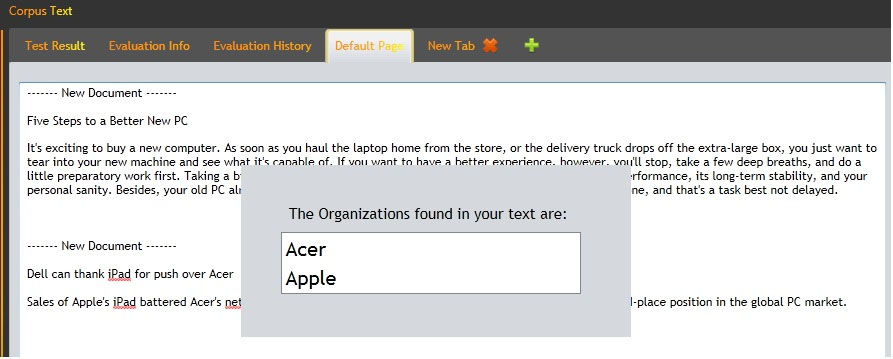}
\end{center}
\caption{Organization names extraction example}
\label{fig:InstanceExtraction1}
\end{figure}

\item extract potential relations between concepts using the syntactic role that words have within sentences;
\item extract potential instances of certain concepts/relations using terms tagged as verbs in the sentence as relation checker(for example,from the sentence, 'John Doe is a great teacher.' we can state that 'John Doe' is an instance of the 'teacher' concept using the fact that the verb 'to be' was discovered and also the fact that term 'teacher' is a concept in the considered ontology;
\item extract instances using lexicon-syntactic patterns in form of regular expressions. This means one or more instances and their related concept are connected by some specific words. These specific words include \textit{'or other, such as, especially, for example'} (e.g. \textit{Laptop producers such as Dell, Toshiba..} );
\end{itemize}

 It is also considered that a user may want to create its own pattern that should be used in retrieving ontology instances from text. For example, a user may need to find all models of a certain car producer. So, he gives the producer's name and specifies that the model should begin either with a capital letter or a number. Many other patterns could be applied in order to find things like prices, dates, person height, camera resolution and so on. For the moment the system tries to create a proof-of-concept and to highlight that ontology population can be automated or at least semi-automated if all the available knowledge and technology are properly used.

\subsection{Ontology Evaluation}
\label{sec:eval}
The Ontology Evaluation component provides methods for evaluating the ontology as a whole or evaluating a specified class from the ontology. The first considered type of evaluation is from the design point of view. This kind of metrics are known as \textit{schema metrics}. 
Metrics in this category indicate the richness, width, depth, and inheritance of an ontology schema design.
The implemented schema metrics are:

\begin{defTemp}Relationship Richness ($RR$) represents the ratio of the number of non-inheritance relationships ($P$), divided by the total number of relationships defined for the ontology, inheritance relationships ($H$) and non-inheritance relationships ($P$).
 \[
    RR = \frac{|P|}{|H|+|P|}
    \]
\end{defTemp}

The $RR$ metric gives information about the diversity of the types of relations in the ontology;

 \begin{defTemp} Inheritance Richness (\textit{IR}) represents the average number of subclasses ($S$) per class ($C$). 
 \[
    IR = \frac{|S|}{|C|}
    \]
\end{defTemp}
IR describes the distribution of information across different levels of the ontology inheritance tree.
This metric distinguishes horizontal ontologies from vertical ontologies.

\begin{defTemp} Attribute Richness(\textit{AR}) counts the average number of attributes ($att$) for each class ($C$) or the average number of properties for each concept in the ontology.
 \[
    AR = \frac{|att|}{|C|}
    \]
\end{defTemp}
$AR$ indicates the amount of information pertaining to instance data.

Ontologies can also be evaluated considering the way data is placed within the ontology or in other words, the amount of real-world knowledge represented by the ontology. 
These metrics are refereed to as knowledge base metrics and include:

 \begin{defTemp}Class Richness($CR$) is the percentage of the number of non-empty classes ($C'$) divided by the total number of classes in the ontology schema ($C$).
 \[
           CR = \frac {|C'|}{|C|}
    \]
\end{defTemp}
This metric is related to how instances are distributed across classes.

\begin{defTemp} Class Connectivity ($Conn(C_i)$) of a class represent the total number of relationships instances that one class has with instances of other classes ($NIREL$).
\[
        Conn(C_i)= |NIREL(C_i)|
     \]
\end{defTemp}
This metric indicates which classes are central in the ontology.

\begin{defTemp}
Class importance ($Imp(C_i$) of a class is defined as the percentage of the number of instances that belong to the inheritance sub-tree rooted at this class ($inst(C_i)$) in the ontology compared to the total number of class instances in the ontology ($CI$).
 \[
    Imp(C_i) = \frac{|Inst(C_i)|}{KB(CI)}
    \]
\end{defTemp}
It helps to identify which areas of the schema are in focus when the instances are added to the ontology.

 \begin{defTemp}Cohesion represents the number of connected components of the graph representing the ontology knowledge base.\end{defTemp}
 Cohesion indicates how well relationships between instances can be traced to discover how two instances are related.

\begin{figure}[h]
\begin{center}
% Requires \usepackage{graphicx}
\includegraphics[width=3.4in]{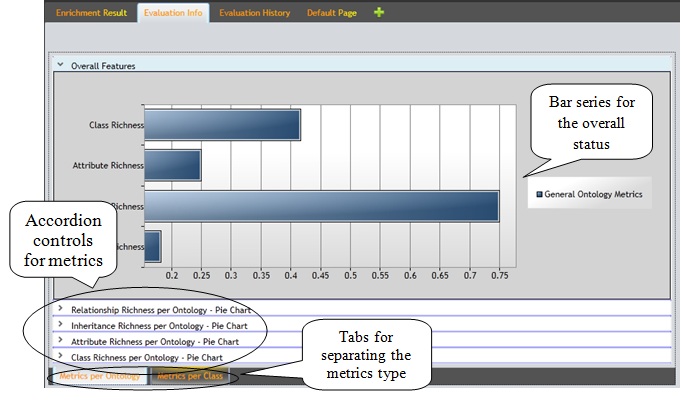}
\end{center}
%\emph{Figure 7: Relationship Richness for 'Wine' ontology}
\caption{Relationship Richness for the 'Wine' ontology.}
\label{fig:GetRelationshipRichness1}
\end{figure}

\begin{figure}[h]
\begin{center}
% Requires \usepackage{graphicx}
\includegraphics[width=3.5in]{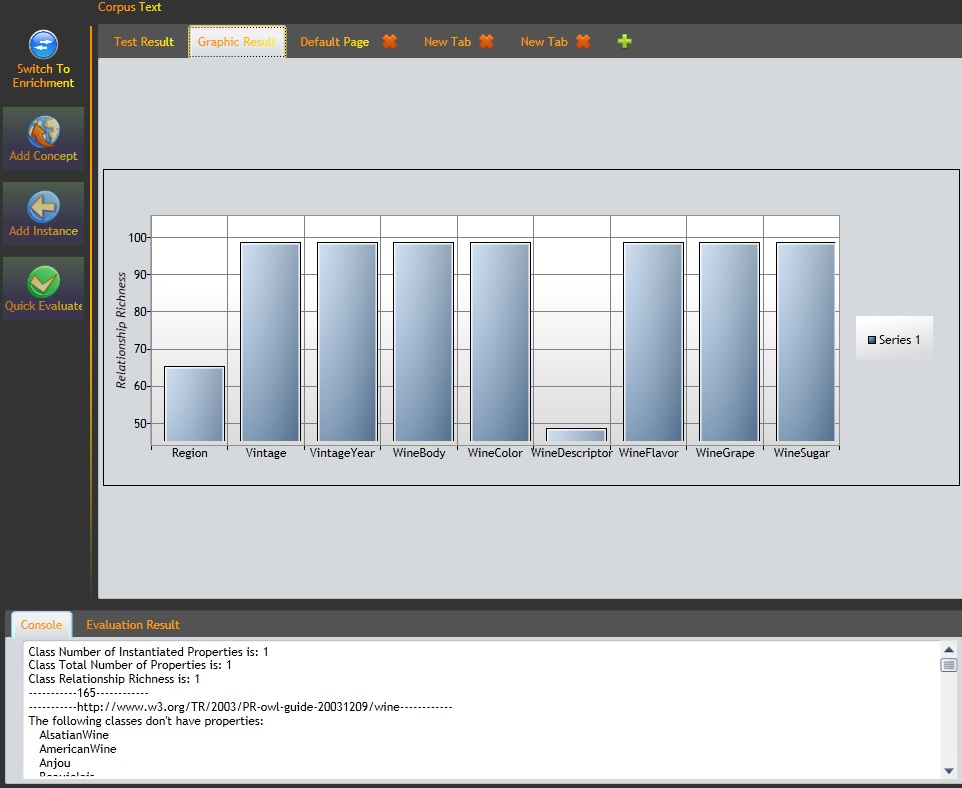}
\end{center}
%\emph{Figure 8: Relationship Richness for 'Wine' ontology after changes to the initial ontology schema}
\caption{Relationship Richness for 'Wine' ontology after changes to the initial ontology.}
\label{fig:GetRelationshipRichness2}

\end{figure}

 Relationship Richness ($RR$) is the percentage of the number of relationships that are being used by instances of the considered class compared to the number of relationships that are defined for the class at the schema level of the ontology.
 Figure~\ref{fig:GetRelationshipRichness1} shows the results obtained for $RR$ on the initial 'Wine' ontology while figure~\ref{fig:GetRelationshipRichness2} illustrates how the $RR$ metric is influenced by changes made to the ontology, after adding new ontology instances and enriching existing instances with new properties in the considered scenario.

\begin{figure}[h]
\begin{center}
% Requires \usepackage{graphicx}
\includegraphics[width=3.4in]{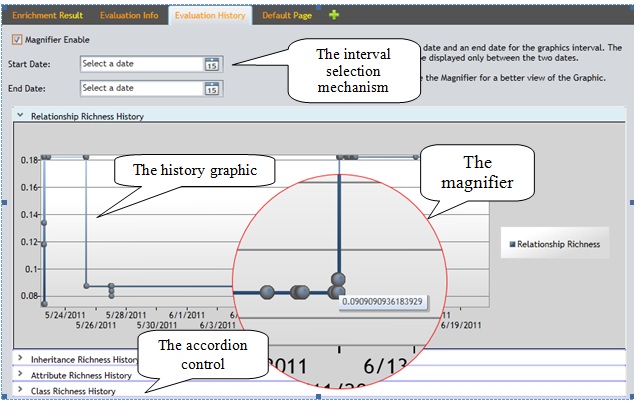}
\end{center}
\caption{History chart of ontology metrics}
\label{fig:HistoryChart1}
\end{figure}

Ontology metrics evolution over time was also an important topic for our proposal. 
The user has the opportunity to store multiple evaluation results on the same ontology and then request for an evaluation chart in order to observe the changes that the ontology has subject to during a certain period (see figure~\ref{fig:HistoryChart1}).

When an ontology is evaluated for several times, the OntoRich system keeps information about the evolution of the ontology from the first time it was loaded by the system. This feature allows to create an evolution-based evaluation by showing how the metrics described above vary in time for an ontology.

Another important feature of the evaluation component is the ability to compare the considered ontology with another ontology from the same domain. The two ontologies are evaluated and the results are presented in a comparative manner so that the user can decide which ontology is better for his own needs.

\section{Testing and Validation}
The considered testing scenario traces a simple IT related ontology through the
process of enrichment provided by the OntoRich system. The interface tree view representation
of the tested ontology can be seen in figure~\ref{fig:MainWindow1}. 
The RSS Reader testing scenario consists in subscribing
to an IT related domain where several RSS feeds from the domain where previously
added. 
To sum up, the following tests have been conducted: 
i) subscribing to an IT related domain using OntoRich RSS Reader appli-
cation; 
ii) creating new text corpus using e-mail content;
iii) extracting new terms using statistical methods illustrated in figure~\ref{fig:TermExtractionResult1};
iv) adding new concepts;
v) extract terms using predefined semantic relations like 'partOf' or 'isKindOf' as figure~\ref{fig:ExtractPartOfRelation1} bears out;
vi) extract terms using semantic hierarchies (in figure~\ref{fig:HyponymTreeComputer1} for a considered term a semantic
hierarchy tree can be obtained by interfacing the WordNet functionality);
vii) instance extraction using NLP facilities (the user can obtain ontology instances using predefined models like Persons, Companies, Dates as depicted in figure~\ref{fig:InstanceExtraction1}).

\begin{figure}
\begin{center}
% Requires \usepackage{graphicx}
\includegraphics[width=3.3in]{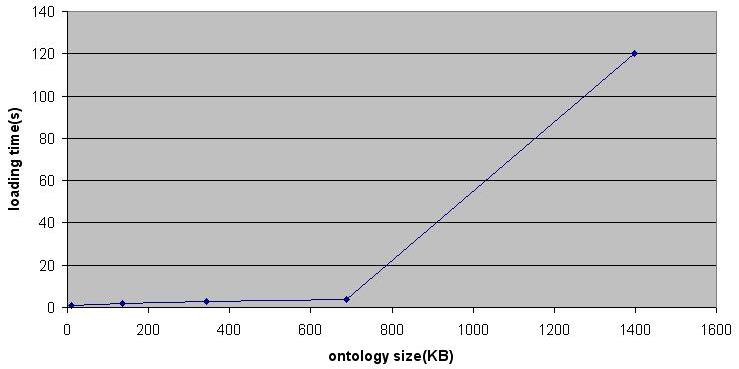}
\end{center}
\caption{System scalability}
\label{fig:results}
\end{figure}

Terms were found using statistical methods and NLP based methods. 
The changed ontology was successfully saved to its original location. The term extraction process took
about 10 seconds because the large amount of text content loaded in the text corpus. This
delay is due to the amount of computation done in order to test each possible term against
the input parameters given by the user (minimum appearance frequency and maximum
number of words in a term). An improvement to this problem could be an approach in
which extracted terms are returned to the user as they are found and not only after the
whole term extraction process completed. Another conclusion was that the application
can scale well for loading ontologies up to 1 MB in size but works harder when the size of
the ontology goes over this threshold (see figure~\ref{fig:results}).

\section{Discussion and Related Work}
\label{sec:related}

In~\cite{GeorgiuGroza11} the ontology is enriched with terms taken from semantic wikis.
In the first step users annotate documents based on the imported ontologies in the system. 
In the second step the initial ontology is enriched based on these documents.
Consequently, new wiki pages would be annotated based on an up-to-date ontology.
The ontology enrichment process is guided by ontology design patterns and heuristics such as the number of annotations based on a concept or an instance.
Differently we use RSS streams guided by Wordnet to extract potential concepts for the given ontology.

The approach for ontology population in~\cite{art1} uses the knowledge available at the online encyclopedia Wikipedia. 
The main tool used is Prot\'eg\'e and the ontology to pe populated was converted to RDF format in order to facilitate further modification. 
%In section 1 of chapter 2 the information source for populating ontologies is presented. 
Wikipedia has a special page that exports articles to XML. 
The analysed scenario automatically exported all the pages of the types of wood that were mentioned on one Wikipedia page. 
As a starting point for building the eventual ontology, an existing taxonomy box on a Wikipedia page is used.
Most of the wood pages have such a taxonomy box, in which a few key concepts are listed with their instances. 
%These classes are, in the case of wood concept, Kingdom, Division, Class, Order, Family and Genus. 
On the page with a list of all the woods, there is a categorization between softwood (conifers) and hardwood (angiosperms). 
This categorization is used together with the one provided by the taxonomy boxes and the extra information provided on some pages about wood use.
From the technical viewpoint, an ontology structure is created in Protege according to the structure of the taxonomy boxes available on the Wikipedia pages. 
In order to extract instances to populate the created ontology a Perl script that replaces Wikipedia tags with equivalent XML tags is used. 
Then another Perl script is used to feed instances to the RDF file corresponding to the created ontology. 
As an evaluation, Protege's built in query tool is used. 
In our approach, the OntoRich system uses RSS feeds as an approach to offer access to structured data on the web, so it is not restricted to a certain number of websites. Practically, every site that offers RSS feeds can be a candidate to the system's repository of domain structured web information.

OntoGenie\cite{Patel03} uses WordNet to convert unstructured data from Web to structured knowledge for Semantic Web. 
Differently, the OntoRich tool makes more advantage of the semantic power provided by WordNet.
The OntoGenie is a semi-automatic tool that takes as input domain ontologies and unstructured data from Web (plain text or HTML), and generates ontology instances (OI) for the given data. 
Similar to our case, the tool uses the linguistic ontology enclosed by WordNet as a bridge between domain ontologies and Web data.
The OntoGenie tool involves a process structured in three main steps:
i) mapping the concepts in a domain ontology into WordNet;
ii) capturing the terms occurring in Web pages; and 
iii) discovering relationships

A comparison between OntoRich and the four major existing systems for ontology enrichment and evaluation Kaon, Neon, OntoQA, ROMEO can be seen in table~\ref{tab:Comparison1}.

\begin{table}[h]
\begin{center}
% Requires \usepackage{graphicx}
\includegraphics[width=3.4in]{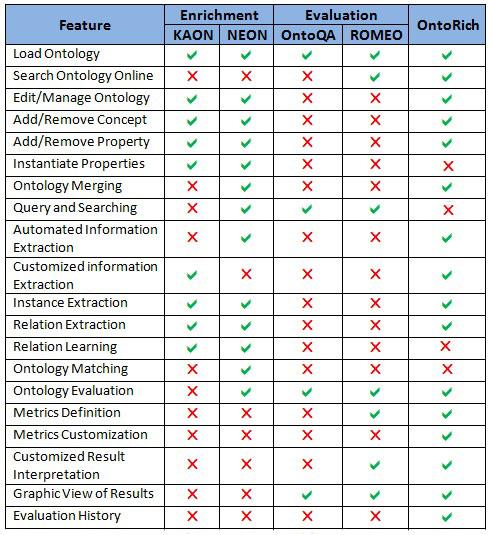}
\end{center}
\caption{Comparison between OntoRich and existing software.}
\label{tab:Comparison1}
\end{table}

KAON (Karlsruhe ontology)~\cite{art3} is an ontology infrastructure, providing ontology learning tools which take non-annotated natural language text as input: TextToOnto (KAON-based) and Text2Onto (KAON2-based). 
Text2Onto is based on the Probabilistic Ontology Model (POM)~\cite{Cimiano05}.
TextToOnto is a tool suite developed to support the ontology engineering process by text mining techniques. 
The usage of the algorithms varies from interactive (the system only makes suggestions) to fully automatic. 
The main features of TextToOnto that were considered when creating OntoRich system are:
\begin{itemize}
\item  Term Extraction - extracts relevant words or terms from a corpus and presents them to the user; the terms can be sorted according to the following measures: Absolute Frequency, TFIDF (term frequency - inverse document frequency), ENTROPY, C-value;
\item  Association Extraction - employs association rules to discover candidate relations between terms in a text corpus;
\item Taxo Builder - creates a concept hierarchy out of the most frequent terms in a corpus or out of the remembered terms by term extraction and adds it to a new ontology model.
\item  Instance Extraction - discovers instances of concepts of a given ontology in a text corpus using patterns.
\item  Relation Learning - discovers candidate relations from text; presents a relationship name to the user as well as a
domain and range for this relationship;
\end{itemize}
OntoRich integrates the OpeNLP library as a support for relation extraction and this approach can increase the probability of finding a correct relationship between two concepts even when the words are not used with their first known sense.

NeOn~\cite{HaaseMS08} is a project involving 14 European partners created with the aim of advancing the state of the art in using ontologies for large-scale semantic applications in the distributed organizations.
The Evolva plugin~\cite{ZablithSdM09} is an ontology evolution tool, which evolves and extends ontologies by identifying new ontological entities from external data sources, and produces a new version of this ontology with the added changes. 
After having built a basic ontology, the ontology engineer can use Evolva to identify and integrate new concepts that arise in the domain during the ontology life cycle.

The main idea considered from the tool proposed by Evolva is the integration of  online ontologies and WordNet to identify links between new concepts and existing concepts in the ontology. Such links are displayed to the user in the form of statements, with the corresponding complete path derived from the source of background knowledge. In our approach we have decided to provide the option of obtaining a taxonomic hierarchy rooted at the specified term from the ontology or from the text corpus.

The most known ontology evaluation frameworks and applications today are OntoQA and ROMEO.
In~\cite{DBLP:conf/semco/TartirA07} the OntoQA tool is presented. 
The authors define the quality of a populated ontology based on a set of schema quality features and knowledge base quality feature (instance based). The Schema Metrics addresses the design of the ontology, while the knowledge base metrics analyze the way data is placed inside the ontology, giving a very good idea about effectiveness, which is very important.

As opposed to the implementation of the OntoQA, OntoRich evaluation component implements all the metrics described there, but in addition allows the user define the importance of each one of them. Due to the fact that an ontology is defining a particular concept from real life, a user usually wants a view (a part) of that concept to be used inside its application. The conclusion of this observation is that the same concept should probably be represented in one way for a kind of application, and in a different way for some other one. The user should be the main arbiter in judging which ontology is best suited to its application.

ROMEO (Requirements-oriented methodology for evaluating ontologies) methodology~\cite{art8} identifies requirements that an ontology is expected to satisfy (or a user is hoping to satisfy), and maps these requirements to some predefined evaluation measures. This approach is very similar with the technique used by OntoRich, except that OntoRich does not impose the user to define the requirements of the desired ontology by itself. OntoRich merely transposes the meaning of the measurements made in logical sentences for an inexperienced user to understand. It just gives the user an extra layer of understanding inside the area of evaluating ontologies, such that he will eventually learn more about ontologies.

As a conclusion, OntoRich combines the two ideas from the above evaluation techniques into an improved technique. It mixes the strongly theoretical part from OntoQA, with the ROMEO methodology of actively involving the user in the process and finally add the idea of allowing the user to make the decision about what ontology to use based on logical facts rather than plain numbers. Logical facts are easier to understand even without strong knowledge in this domain. A simple, but yet efficient ontology evaluation method, that integrates a user friendly interface will hopefully make this domain more accessible to normal users who just need the best ontology for their application.
%~\cite{Solskinnsbakk2012} presents an approach to evaluate the quality of hierarchical relations in ontologies and folksonomy based structures. 

\section{Conclusions}
\label{sec:conclusions}

In this paper the main idea presented is that of using together a set of tools and methods already known in the domain of Semantic Web in order to create a powerful tool for both ontology enrichment and evaluation.
An RSS Reader is the considered automatic web content extraction method. 
RSS feeds are an important source of information as they provide constantly updated web content. 
New instances of some already existing ontology are easily found within the content of domain specialized RSS feeds.
In order to extract new concepts, relationships and instances for an ontology statistical methods as term frequency or TF-IDF (term frequency - inverse document frequency) were used. RiTa Wordnet API and OpenNLP API provided also an important backup. The WordNet ontology is used
in order to examine and extract candidates for ontology enrichment taking advantage of various features such as word stems, word hyponyms or word meronyms. With integration of the OpenNLP API the system gained access to syntactical analysis of a text, so sentence splitting and part-of-speech tagging were added as features in order to improve the quality of discovered terms in relation to the context where they appeared.

Ontology evaluation was also an objective, so options for evaluating the ontology from the design point of view and also from the knowledge base perspective were added to the OntoRich system. 
Metrics for evaluating the entire ontology schema or for evaluating a specific classes from the ontology are implemented. 
Comparative evaluation of the new ontology against the old one is also presented to facilitate the quality assessment of an ontology.

Ongoing work regards refinement of the ontology population algorithms and evaluation components. 
WordNet ontology can be exploited even more, and with the help of OpenNLP, relationships between concepts from the ontology or new domain concepts could be discovered even when the context of use causes word ambiguity. 
Information extraction using Google web services and DMOZ URL extractor will be a point of interest in improving the quality of retrieved web content. 
Pattern-based approach for extracting concepts and instances from a text corpus is also something worth to be
taken into consideration in the near future. 
This method will provide the user to describe exactly the type of information that he is looking for in the text.
In the ontology evaluation field OntoRich will address logical and rule-based approaches for ontology validation and quality evaluation. 
With the integration of argumentation theory~\cite{GrozaIndrie11} we are extending Ontorich to provide support for collaborative distributed ontology enrichment. 
%This methods use rules which are built in the ontology languages and rules that users provide in order to detect conflicts in ontologies.

\section*{Acknowledgment}
We are grateful to the anonymous reviewers for their useful comments. 
The work has been co-funded by the Sectoral Operational Programme Human Resources Development 2007-2013 of the Romanian Ministry of 
Labour, Family and Social Protection through the Financial Agreement POSDRU/89/1.5/S/62557.

\bibliographystyle{IEEEtran}
\bibliography{oe}

% that's all folks
\end{document}